\documentclass[11pt,a4paper]{article}
\usepackage[hyperref]{acl2019}
\usepackage{times}
\usepackage{latexsym}

\usepackage{url}

\usepackage{graphicx}
\usepackage{amsfonts}
\usepackage{amsmath}

\usepackage{tikz}
\usetikzlibrary{arrows,shapes,trees}
\usepackage{amssymb}
\usepackage{enumitem}

\aclfinalcopy % Uncomment this line for the final submission
\title{HIBERT: Document Level Pre-training of Hierarchical Bidirectional Transformers for Document Summarization}

\author{Xingxing Zhang, Furu Wei \and Ming Zhou \\
	Microsoft Research Asia, Beijing, China \\
	\texttt{\{xizhang,fuwei,mingzhou\}@microsoft.com} 
}
\date{}

\begin{document}
\maketitle
\begin{abstract}
Neural extractive summarization models usually employ a hierarchical encoder for document encoding and they are trained using sentence-level labels, which are created heuristically using rule-based methods. Training the hierarchical encoder with these \emph{inaccurate} labels is challenging. Inspired by the recent work on pre-training transformer sentence encoders \cite{devlin:2018:arxiv}, we propose {\sc Hibert} (as shorthand for {\bf HI}erachical {\bf B}idirectional {\bf E}ncoder {\bf R}epresentations from {\bf T}ransformers) for document encoding and a method to pre-train it using unlabeled data. We apply the pre-trained {\sc Hibert} to our summarization model and it outperforms its randomly initialized counterpart by 1.25 ROUGE on the CNN/Dailymail dataset and by 2.0 ROUGE on a version of New York Times dataset. We also achieve the state-of-the-art performance on these two datasets.
\end{abstract}

\section{Introduction}

% Mainly talk about summarization
Automatic document summarization is the task of rewriting a document into its shorter form while still retaining its important content. Over the years, many paradigms for document summarization have been explored (see \newcite{Nenkova:McKeown:2011} for an overview). The most popular two among them are \emph{extractive} approaches and \emph{abstractive} approaches. As the name implies, \emph{extractive} approaches generate summaries by \emph{extracting} parts of the original document (usually sentences), while \emph{abstractive} methods may generate new words or phrases which are not in the original document. 

Extractive summarization is usually modeled as a sentence ranking problem with length constraints (e.g., max number of words or sentences). Top ranked sentences (under constraints) are selected as summaries. Early attempts mostly leverage manually engineered features \cite{filatova:2004:acl:workshop}. Based on these sparse features, sentence are selected using a classifier or a regression model. Later, the feature engineering part in this paradigm is replaced with neural networks. \newcite{cheng:2016:acl} propose a hierarchical long short-term memory network (LSTM; \citealt{hochreiter:1997:nc}) to encode a document and then use another LSTM to predict binary labels for each sentence in the document. This architecture is widely adopted  recently \cite{nallapati:2017:aaai,Narayan:ea:2018,zhang:2018:emnlp}. Our model also employs a hierarchical document encoder, but we adopt a hierarchical transformer \cite{vaswani:2017:nips} rather a hierarchical LSTM. Because recent studies \cite{vaswani:2017:nips,devlin:2018:arxiv} show the transformer model performs better than LSTM in many tasks.
% feature engineering \cite{filatova:2004:acl:workshop} and scores used for ranking come from probabilities of positive class in binary classifiers \cite{kupiec:1995:sigir}, 

Abstractive models do not attract much attention until recently. They are mostly based on sequence to sequence (seq2seq) models \cite{bahdanau:2015:iclr}, where a document is viewed a sequence and its summary is viewed as another sequence. Although seq2seq based summarizers can be equipped with copy mechanism \cite{gu:2016:acl,see:2017:acl}, coverage model \cite{see:2017:acl} and reinforcement learning \cite{paulus:2017:arxiv}, there is still no guarantee that the generated summaries are grammatical and convey the same meaning as the original document does. It seems that extractive models are more reliable than their abstractive counterparts.

However, extractive models require sentence level labels, which are usually not included in most summarization datasets (most datasets only contain document-summary pairs). Sentence labels are usually obtained by rule-based methods (e.g., maximizing the ROUGE score between a set of sentences and reference summaries) and may not be accurate. Extractive models proposed recently \cite{cheng:2016:acl,nallapati:2017:aaai} employ hierarchical document encoders and even have neural decoders, which are complex. Training such complex neural models with \emph{inaccurate} \emph{binary} labels is challenging. We observed in our initial experiments on one of our dataset that our extractive model (see Section \ref{sec:sum} for details) overfits to the training set quickly after the second epoch, which indicates the training set may not be fully utilized. Inspired by the recent pre-training work in natural language processing \cite{peters:2018:naacl,radford:2018:nips,devlin:2018:arxiv}, our solution to this problem is to first pre-train the ``complex''' part (i.e., the hierarchical  encoder) of the extractive model on unlabeled data and then we learn to classify sentences with our model initialized from the pre-trained encoder.  In this paper, we propose \mbox{{\sc Hibert}}, which stands for {\bf HI}erachical {\bf B}idirectional {\bf E}ncoder {\bf R}epresentations from {\bf T}ransformers. We design an unsupervised method to pre-train \mbox{{\sc Hibert}} for document modeling. We apply the pre-trained \mbox{{\sc Hibert}} to the task of document summarization and achieve state-of-the-art performance on both the CNN/Dailymail and New York Times dataset. % We also outperforms several recent models in human evaluation. % This pre-training and fine-tuning  paradigm is similar with that of ELMo \cite{peters:2018:naacl}, OpenAI GPT \cite{radford:2018:nips} and BERT \cite{devlin:2018:arxiv}. The main difference is that we aim to learn the representation of 

\section{Related Work}

In this section, we introduce work on extractive summarization, abstractive summarization and pre-trained natural language processing models. For a more comprehensive review of summarization, we refer the interested readers to \newcite{Nenkova:McKeown:2011} and \newcite{Mani:01}.
% Several parts: summarization, model pretraining 
\paragraph{Extractive Summarization} Extractive summarization aims to select important sentences (sometimes other textual units such as elementary discourse units (EDUs)) from a document as its summary. It is usually modeled as a sentence ranking problem by using the scores from classifiers \cite{kupiec:1995:sigir}, sequential labeling models \cite{conroy:2001:sigir} as well as integer linear programmers \cite{woodsend:2010:acl}. Early work with these models above mostly leverage human engineered features such as sentence position and length \cite{radev:2004}, word frequency \cite{nenkova:2006:sigir} and event features \cite{filatova2004event}. 

As the very successful applications of neural networks to a wide range of NLP tasks, the manually engineered features (for document encoding) are replaced with hierarchical LSTMs/CNNs and the sequence labeling (or classification) model is replaced with an LSTM decoder \cite{cheng:2016:acl,nallapati:2017:aaai}. The architecture is widely adopted in recent neural extractive models and is extended with reinforcement learning \cite{Narayan:ea:2018,dong:2018:emnlp}, latent variable models \cite{zhang:2018:emnlp}, joint scoring \cite{zhou:2018:acl} and iterative document representation \cite{chen:2018:emnlp}. Recently, transformer networks \cite{vaswani:2017:nips} achieves good performance in machine translation \cite{vaswani:2017:nips} and a range of NLP tasks \cite{devlin:2018:arxiv,radford:2018:nips}. Different from the extractive models above, we adopt a hierarchical Transformer for document encoding and also propose a method to pre-train the document encoder.

% Early attempts mostly leverage manually engineered features \cite{filatova:2004:acl:workshop}. Based on these features, sentence are selected using binary classifiers \cite{kupiec:1995:sigir}, hidden markov models \cite{conroy:2001:sigir}, graph based models \cite{mihalcea:2005:acl} and integer linear programming \cite{woodsend:2010:acl}. 

\paragraph{Abstractive Summarization}  Abstractive summarization aims to generate the summary of a document with rewriting. Most recent abstractive models \cite{nallapati:2016:arxiv} are based on neural sequence to sequence learning \cite{bahdanau:2015:iclr,sutskever:2014:nips}. However, the generated summaries of these models can not be controlled (i.e., their meanings can be quite different from the original and contents can be repeated). Therefore, copy mechanism \cite{gu:2016:acl}, coverage model \cite{see:2017:acl} and reinforcement learning model optimizing ROUGE \cite{paulus:2017:arxiv} are introduced. These problems are alleviated but not solved. There is also an interesting line of work combining extractive and abstractive summarization with reinforcement learning \cite{chen:2018:acl}, fused attention \cite{hsu:2018:acl} and bottom-up attention \cite{gehrmann:2018:emnlp}. Our model, which is a very good extractive model, can be used as the sentence extraction component in these models and potentially improves their performance. 

\paragraph{Pre-trained NLP Models} Most model pre-training methods in NLP leverage the natural ordering of text. For example, {\tt word2vec} uses the surrounding words within a fixed size window to predict the word in the middle with a log bilinear model. The resulting word embedding table can be used in other downstream tasks. There are other word embedding pre-training methods using similar techniques \cite{pennington:2014:emnlp,bojanowski:2017:tacl}. \newcite{peters:2018:naacl} and \newcite{radford:2018:nips} find even a sentence encoder (not just word embeddings) can also be pre-trained with language model objectives (i.e., predicting the next or previous word). Language model objective is unidirectional, while many tasks can leverage the context in both directions. Therefore, \newcite{devlin:2018:arxiv} propose the naturally bidirectional masked language model objective (i.e., masking several words with a special token in a sentence and then predicting them). All the methods above aim to pre-train word embeddings or sentence encoders, while our method aims to pre-train the hierarchical document encoders (i.e., hierarchical transformers), which is important in summarization. 

\section{Model}
In this section, we present our model {\sc Hibert}.
% which stands for {\bf H}ierarchical {\bf B}idirectional {\bf E}ncoder {\bf R}epresentations from {\bf T}ransformers\footnote{We name our model {\sc Hibert}  after BERT \cite{devlin:2018:arxiv}, which inspired us.}. 
We first introduce how documents are represented in {\sc Hibert}. We then describe our method to pre-train {\sc Hibert} and finally move on to the application of {\sc Hibert} to summarization.

\begin{figure}[t]
	\centering
	\includegraphics[width=0.5\textwidth]{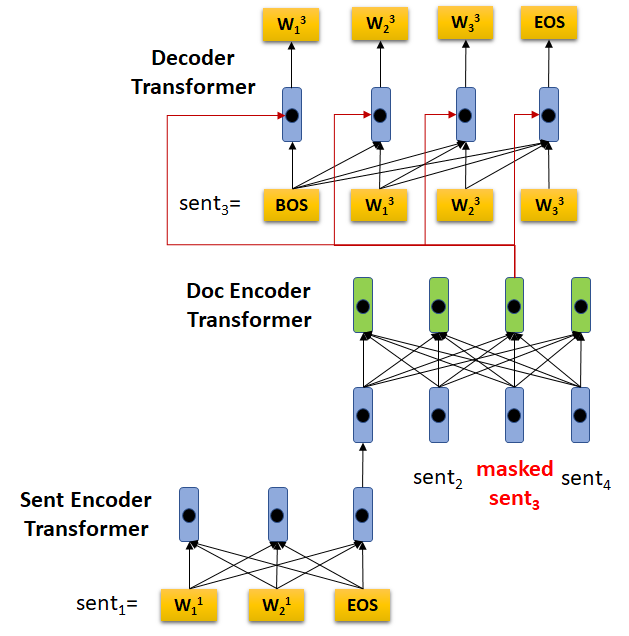}
	\caption{The architecture of {\sc Hibert} during training. $\text{sent}_i$ is a sentence in the document above, which has four sentences in total. $\text{sent}_3$ is masked during encoding and the decoder predicts the original $\text{sent}_3$.}
	\label{fig:hbert}
\end{figure}

\subsection{Document Representation}
\label{sec:doc_repr}
Let $\mathcal{D} = (S_1, S_2, \dots, S_{| \mathcal{D} |})$ denote a document,  where $S_i = (w_1^i, w_2^i, \dots, w_{|S_i|}^i)$ is a sentence in $\mathcal{D}$ and $w_j^i$ a word in $S_i$. Note that following common practice in natural language processing literatures, $w_{|S_i|}^i$ is an artificial {\tt EOS} (End Of Sentence) token. To obtain the representation of $\mathcal{D}$, we use two encoders: a \emph{sentence encoder} to transform each sentence in $\mathcal{D}$ to a vector and a \emph{document encoder} to learn sentence representations given their surrounding sentences as context. Both the \emph{sentence encoder} and \emph{document encoder} are based on the Transformer encoder described in \newcite{vaswani:2017:nips}. As shown in Figure \ref{fig:hbert}, they are nested in a hierarchical fashion. A transformer encoder usually has multiple layers and each layer is composed of a multi-head self attentive sub-layer followed by a feed-forward sub-layer with residual connections \cite{he:2016:cvpr} and layer normalizations \cite{ba:2016:arxiv}. For more details of the Transformer encoder, we refer the interested readers to \newcite{vaswani:2017:nips}. To learn the representation of $S_i$, $S_i= (w_1^i, w_2^i, \dots, w_{|S_i|}^i)$ is first mapped into continuous space
%\begin{equation}
%\label{eq:emb}
%\mathbf{E}_i = (e_1^i+q_1, e_2^i+q_2, \dots, e_{|S_i|}^i+q_{|S_i|})
%\end{equation}
\begin{align}
\label{eq:emb}
\begin{split}
\mathbf{E}_i = (\mathbf{e}_1^i, \mathbf{e}_2^i, \dots, \mathbf{e}_{|S_i|}^i) \\
\quad  \quad  \text{where} \quad \mathbf{e}_j^i = e(w_j^i) + \mathbf{p}_j
\end{split}
\end{align}
where $e(w_j^i)$ and $\mathbf{p}_j$ are the word and positional embeddings of $w_j^i$, respectively. The word embedding matrix is randomly initialized and we adopt the sine-cosine positional embedding \cite{vaswani:2017:nips}\footnote{We use the sine-cosine embedding because it works well and do not introduce additional trainable parameters.}. Then the \emph{sentence encoder} (a Transformer) transforms $\mathbf{E}_i$ into a list of hidden representations $(\mathbf{h}_1^i, \mathbf{h}_2^i, \dots, \mathbf{h}_{|S_i|}^i)$. We take the last hidden representation $\mathbf{h}_{|S_i|}^i$ (i.e., the representation at the {\tt EOS} token) as the representation of sentence $S_i$. Similar to the representation of each word in $S_i$, we also take the sentence position into account. The final representation of $S_i$ is 
\begin{equation}
\hat{\mathbf{h}}_i = \mathbf{h}_{|S_i|}^i + \mathbf{p}_i
\end{equation}
Note that words and sentences share the same positional embedding matrix. 

In analogy to the \emph{sentence encoder}, as shown in Figure \ref{fig:hbert}, the \emph{document encoder} is yet another Transformer but applies on the sentence level. After running the Transformer on a sequence of sentence representations $( \hat{\mathbf{h}}_1, \hat{\mathbf{h}}_2, \dots, \hat{\mathbf{h}}_{|\mathcal{D}|} )$, we obtain the context sensitive sentence representations $( \mathbf{d}_1, \mathbf{d}_2, \dots, \mathbf{d}_{|\mathcal{D}|} )$. Now we have finished the encoding of a document with a hierarchical bidirectional transformer encoder {\sc Hibert}. Note that in previous work, document representation are also learned with hierarchical models, but each hierarchy is a Recurrent Neural Network \cite{nallapati:2017:aaai,zhou:2018:acl} or Convolutional Neural Network \cite{cheng:2016:acl}. We choose the Transformer because it outperforms CNN and RNN in machine translation \cite{vaswani:2017:nips}, semantic role labeling \cite{strubell:2018:emnlp} and other NLP tasks \cite{devlin:2018:arxiv}. In the next section we will introduce how we train {\sc Hibert} with an unsupervised training objective.

\subsection{Pre-training}
\label{sec:pretrain}
Most recent encoding neural models used in NLP (e.g., RNNs, CNNs or Transformers) can be pre-trained by  predicting a word in a sentence (or a text span) using other words within the same sentence (or span). For example, ELMo \cite{peters:2018:naacl} and OpenAI-GPT \cite{radford:2018:nips} predict a word using all words on its left (or right); while word2vec \cite{mikolov:2013:nips} predicts one word with its surrounding words in a fixed window and BERT \cite{devlin:2018:arxiv} predicts (masked) missing words in a sentence given all the other words.

All the models above learn the representation of a sentence, where its basic units are words. \mbox{{\sc Hibert}} aims to learn the representation of a document, where its basic units are sentences. Therefore, a natural way of pre-training a document level model (e.g., {\sc Hibert}) is to predict a sentence (or sentences) instead of a word (or words). We could predict a sentence in a document with all the sentences on its left (or right) as in a (document level) language model. However, in summarization, context on both directions are available. We therefore opt to predict a sentence using all sentences on both its left and right. 

\paragraph{Document Masking} Specifically, suppose $\mathcal{D} = (S_1, S_2, \dots, S_{| \mathcal{D} |})$ is a document,  where $S_i = (w_1^i, w_2^i, \dots, w_{|S_i|}^i)$ is a sentence in it. We randomly select 15\% of the sentences in $\mathcal{D}$ and mask them. Then, we predict these masked sentences. The prediction task here is similar with the \emph{Cloze} task \cite{taylor:1953:sage,devlin:2018:arxiv}, but the missing part is a sentence. However,  during test time the input document is not masked, to make our model can adapt to documents without masks, we do not always mask the selected sentences. Once a sentence is selected (as one of the 15\% selected masked sentences), we transform it with one of three methods below. We will use an example to demonstrate the transformation. For instance, we have the following document and the second sentence is selected\footnote{There might be multiple sentences selected in a document, but in this example there is only one.}:

{\tt \noindent William Shakespeare is a poet . He died in 1616 . He is regarded as the greatest writer .}

In 80\% of the cases, we mask the selected sentence (i.e., we replace each word in the sentence with a mask token {\tt [MASK]}). The document above becomes {\tt William Shakespeare is a poet . [MASK] [MASK] [MASK] [MASK] [MASK] He is regarded as the greatest writer .} (where ``{\tt He died in 1616 . }'' is masked). 

In 10\% of the cases,  we keep the selected sentence as it is. This strategy is to simulate the input document during test time (with no masked sentences).

In the rest 10\% cases, we replace the selected sentence with a random sentence. In this case, the document after transformation is {\tt William Shakespeare is a poet . Birds can fly . He is regarded as the greatest writer .}  The second sentence is replaced with ``{\tt Birds can fly .}'' This strategy intends to add some noise during training and make the model more robust.

% two sections/paragraphs: 1. document masking; 2. sentence prediction.  Use the term mask
\paragraph{Sentence Prediction} After the application of the above procedures to a document $\mathcal{D} = (S_1, S_2, \dots, S_{| \mathcal{D} |})$, we obtain the masked document $\widetilde{ \mathcal{D} }= (\tilde{S_1}, \tilde{S_2}, \dots, \tilde{S_{| \mathcal{D} |}})$. Let $\mathcal{K} $ denote the set of indicies of selected sentences in $\mathcal{D}$. Now we are ready to predict the masked sentences $\mathcal{M} =  \{S_k | k \in \mathcal{K} \}$ using $\widetilde{ \mathcal{D} }$. 
% As shown in Figure \ref{fig:hbert}, we use a Transformer decoder to prediction the masked sentences. 
We first apply the hierarchical encoder {\sc Hibert} in Section \ref{sec:doc_repr} to $\widetilde{ \mathcal{D} }$ and obtain its context sensitive sentence representations $( \tilde{ \mathbf{d}_1 }, \tilde{ \mathbf{d}_2 }, \dots, \tilde{ \mathbf{d}_{| \mathcal{D} |} } )$. We will demonstrate how we predict the masked sentence $S_k = (w_0^k, w_1^k, w_2^k, \dots, w_{|S_k|}^k)$ one word per step ($w_0^k$ is an artificially added {\tt BOS} token). At the $j$th step, we predict $w_j^k$ given $w_0^k,\dots,w_{j-1}^k$ and $\widetilde{ \mathcal{D} }$. $\tilde{ \mathbf{d}_k }$ already encodes the information of $\widetilde{ \mathcal{D} }$ with a focus around its $k$th sentence $\tilde{S_k}$. As shown in Figure \ref{fig:hbert}, we employ a Transformer decoder \cite{vaswani:2017:nips} to predict $w_j^k$ with $\tilde{ \mathbf{d}_k }$ as its additional input. The transformer decoder we used here is slightly different from the original one. The original decoder employs two multi-head attention layers to include both the context in encoder and decoder, while we only need one to learn the decoder context, since the context in encoder is a vector (i.e., $\tilde{ \mathbf{d}_k }$). Specifically, after applying the word and positional embeddings to ($w_0^k,\dots,w_{j-1}^k$), we obtain $\widetilde{ \mathbf{E} }^k_{1:j-1} = (\tilde{\mathbf{e}_0^k}, \dots, \tilde{\mathbf{e}_{j-1}^k})$ (also see Equation \ref{eq:emb}). Then we apply multi-head attention sub-layer to $\widetilde{ \mathbf{E} }^k_{1:j-1}$:
\begin{align}
\label{eq:mhead}
\begin{split}
\tilde{\mathbf{h}_{j-1}} &= \text{MultiHead}(\mathbf{q}_{j-1}, \mathbf{K}_{j-1}, \mathbf{V}_{j-1}) \\
\mathbf{q}_{j-1} &= \mathbf{W}^Q \: \tilde{\mathbf{e}_{j-1}^k}  \\
\mathbf{K}_{j-1} &= \mathbf{W}^K \: \widetilde{ \mathbf{E} }^k_{1:j-1} \\
\mathbf{K}_{j-1} &= \mathbf{W}^V \: \widetilde{ \mathbf{E} }^k_{1:j-1}
\end{split}
\end{align}
where $\mathbf{q}_{j-1}$, $\mathbf{K}_{j-1}$, $\mathbf{V}_{j-1}$ are the input query, key and value matrices of the multi-head attention function \cite{vaswani:2017:nips} $\text{MultiHead}(\cdot, \cdot, \cdot)$, respectively. $\mathbf{W}^Q \in \mathbb{R}^{d \times d}$, $\mathbf{W}^K \in \mathbb{R}^{d \times d}$ and $\mathbf{W}^V \in \mathbb{R}^{d \times d}$ are weight matrices. 

Then we include the information of $\widetilde{ \mathcal{D} }$ by addition:
\begin{equation}
\tilde{\mathbf{x}_{j-1}} = \tilde{\mathbf{h}_{j-1}} + \tilde{ \mathbf{d}_k }
\end{equation}
% Similar strategy has been adopted in \newcite{kiros:2015:nips}.
We also follow a feedforward sub-layer (one hidden layer with ReLU \cite{glorot:2011:aistats} activation function) after $\tilde{\mathbf{x}_{j-1}}$ as in \newcite{vaswani:2017:nips}:
\begin{equation}
\label{eq:ff}
\tilde{\mathbf{g}_{j-1}} = \mathbf{W}^{ff}_2 \max(0, \mathbf{W}^{ff}_1 \tilde{\mathbf{x}_{j-1}} + \mathbf{b}_1) + \mathbf{b}_2
\end{equation}
Note that the transformer decoder can have multiple layers by applying Equation (\ref{eq:mhead}) to (\ref{eq:ff}) multiple times and we only show the computation of one layer for simplicity.
% and obtain $\tilde{\mathbf{g}_{j-1}}$. 
% Note that the transformer decoder can have multiple layers and we only show 

The probability of $w_j^k$ given $w_0^k,\dots,w_{j-1}^k$ and $\widetilde{ \mathcal{D} }$ is:
\begin{equation}
p( w_j^k | w_{0:j-1}^k, \widetilde{ \mathcal{D} } ) = \text{softmax}( \mathbf{W}^O \: \tilde{\mathbf{g}_{j-1}} )
\end{equation}
Finally the probability of all masked sentences $ \mathcal{M} $ given $\widetilde{ \mathcal{D} }$ is 
\begin{equation}
p(\mathcal{M} | \widetilde{ \mathcal{D} }) = \prod_{k \in \mathcal{K}} \prod_{j=1}^{|S_k|} p(w_j^k | w_{0:j-1}^k, \widetilde{ \mathcal{D} })
\end{equation}
The model above can be trained by minimizing the negative log-likelihood of all masked sentences given their paired documents. We can in theory have unlimited amount of training data for \mbox{{\sc Hibert}}, since they can be generated automatically from (unlabeled) documents. Therefore, we can first train {\sc Hibert} on large amount of data and then apply it to downstream tasks. In the next section, we will introduce its application to document summarization. 

\subsection{Extractive Summarization}
\label{sec:sum}
\begin{figure}[t]
	\centering
	\includegraphics[width=0.5\textwidth]{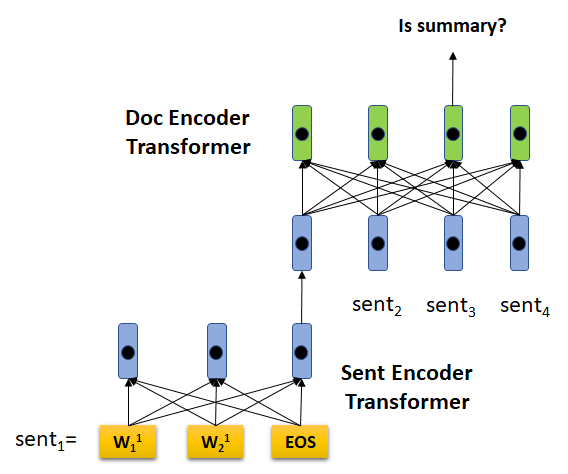}
	\caption{The architecture of our extractive summarization model. The sentence and document level transformers can be pretrained.}
	\label{fig:sum}
\end{figure}

% Summarization is the task of rewriting a document into its shorter version while still retaining its most important content. 
Extractive summarization selects the most important sentences in a document as its summary. In this section, summarization is modeled as a sequence labeling problem. Specifically, a document is viewed as a sequence of sentences and a summarization model is expected to assign a {\tt True} or {\tt False} label for each sentence, where {\tt True} means this sentence should be included in the summary. In the following, we will introduce the details of our summarization model based \mbox{{\sc Hibert}}.

Let $\mathcal{D} = (S_1, S_2, \dots, S_{| \mathcal{D} |})$ denote a document and $Y = (y_1, y_2, \dots, y_{| \mathcal{D} |})$ its sentence labels (methods for obtaining these labels are in Section \ref{sec:dataset}). 
As shown in Figure \ref{fig:sum}, we first apply the hierarchical bidirectional transformer encoder \mbox{{\sc Hibert}} to $\mathcal{D}$ and yields the context dependent representations for all sentences $( \mathbf{d}_1, \mathbf{d}_2, \dots, \mathbf{d}_{|\mathcal{D}|} )$. The probability of the label of $S_i$ can be estimated using an additional linear projection and a softmax:
\begin{equation}
\label{eq:sum}
p( y_i | \mathcal{D} ) = \text{softmax}(\mathbf{W}^S \: \mathbf{d}_i)
\end{equation}
where $\mathbf{W}^S \in \mathbb{R}^{2 \times d}$. The summarization model can be trained by minimizing the negative log-likelihood of all sentence labels given their paired documents.
% We can pre-train the hierarchical transformer encoder with the objective in Section \ref{sec:pretrain} and then initialize and finetune the summarization model on a summarization dataset.

\section{Experiments}
In this section we assess the performance of our model on the document summarization task. We first introduce the dataset we used for pre-training and the summarization task and give implementation details of our model. We also compare our model against multiple previous models.

\subsection{Datasets}
\label{sec:dataset}
We conducted our summarization experiments on the non-anonymous version CNN/Dailymail (\mbox{CNNDM}) dataset \cite{hermann:2015:nips,see:2017:acl}, and the New York Times dataset \cite{durrett:2016:acl,xu:2019:arxiv}. For the CNNDM dataset, we preprocessed the dataset using the scripts from the authors of \newcite{see:2017:acl}\footnote{Scripts publicly available at {\url{https://github.com/abisee/cnn-dailymail}} }. The resulting dataset contains 287,226 documents with summaries for training, 13,368 for validation and 11,490 for test. 
% We also access the performance of our models on the DUC2002 single document summarization task. The DUC2002 dataset contains 567 documents, which belongs to 59 news topics. Each document is paired with two human written summaries. 
Following \cite{xu:2019:arxiv,durrett:2016:acl}, we created the NYT50 dataset by removing the documents whose summaries are shorter than 50 words from New York Times dataset. We used the same training/validation/test splits as in \newcite{xu:2019:arxiv}, which contain 137,778 documents for training, 17,222 for validation and 17,223 for test.
To create sentence level labels for extractive summarization,
we used a strategy similar to \newcite{nallapati:2017:aaai}. We label
the subset of sentences in a document that maximizes \textsc{Rouge} \cite{lin:2004:acl:w}
(against the human summary) as {\tt True} and all other sentences as
{\tt False}.

To unsupervisedly pre-train our document model {\sc Hibert} (see Section \ref{sec:pretrain} for details), we created the GIGA-CM dataset (totally 6,626,842 documents and 2,854 million words), which includes 6,339,616 documents sampled from the English Gigaword\footnote{https://catalog.ldc.upenn.edu/LDC2012T21} dataset and the training split of the \mbox{CNNDM} dataset. We used the validation set of CNNDM as the validation set of GIGA-CM as well. As in \newcite{see:2017:acl}, documents and summaries in CNNDM, NYT50 and GIGA-CM are all segmented and tokenized using Stanford CoreNLP toolkit \cite{manning:2014:acldemo}.  To reduce the vocabulary size, we applied byte pair encoding (BPE; \citealt{sennrich:2016:acl}) to all of our datasets. To limit the memory consumption during training, we limit the length of each sentence to be 50 words (51th word and onwards are removed) and split documents with more than 30 sentences into smaller documents with each containing at most 30 sentences. 

\subsection{Implementation Details}
Our model is trained in three stages, which includes two pre-training stages and one finetuning stage. The first stage is the open-domain pre-training and in this stage we train {\sc Hibert} with the pre-training objective (Section \ref{sec:pretrain}) on GIGA-CM dataset. In the second stage, we perform the in-domain pre-training on the CNNDM (or NYT50) dataset still with the same pre-training objective. In the final stage, we finetune {\sc Hibert} in the summarization model (Section \ref{sec:sum}) to predict extractive sentence labels on CNNDM (or NYT50).

The sizes of the sentence and document level Transformers as well as the Transformer decoder in {\sc Hibert} are the same. Let $L$ denote the number of layers in Transformer, $H$ the hidden size and $A$ the number of attention heads. As in \cite{vaswani:2017:nips,devlin:2018:arxiv}, the hidden size of the feedforward sublayer is $4H$. We mainly trained two model sizes: $\text{\sc Hibert}_S$ ($L=6$, $H=512$ and $A=8$) and $\text{\sc Hibert}_M$ ($L=6$, $H=768$ and $A=12$). We trained both $\text{\sc Hibert}_S$ and $\text{\sc Hibert}_M$ on a single machine with 8 Nvidia Tesla V100 GPUs with a batch size of 256 documents. We optimized our models using Adam with learning rate of 1e-4, $\beta_1=0.9$, $\beta_2=0.999$, L2 norm of 0.01, learning rate warmup 10,000 steps and learning rate decay afterwards using the strategies in \newcite{vaswani:2017:nips}. The dropout rate in all layers are 0.1. In pre-training stages, we trained our models until validation perplexities do not decrease significantly (around 45 epochs on GIGA-CM dataset and 100 to 200 epochs on CNNDM and NYT50). Training $\text{\sc Hibert}_M$ for one epoch on GIGA-CM dataset takes approximately 20 hours. 

Our models during fine-tuning stage can be trained on a single GPU. The hyper-parameters are almost identical to these in the pre-training stages except that the learning rate is 5e-5, the batch size is 32, the warmup steps are 4,000 and we train our models for 5 epochs. During inference, we rank sentences using 
$p( y_i | \mathcal{D} ) $ (Equation (\ref{eq:sum})) and choose the top $K$ sentences as summary, where $K$ is tuned on the validation set.

\subsection{Evaluations}
We evaluated the quality of summaries from different systems automatically using ROUGE \cite{lin:2004:acl:w}. We reported the full length F1 based ROUGE-1, ROUGE-2 and ROUGE-L on the \mbox{CNNDM} and NYT50 datasets. 
% The limited length recall based ROUGE-1, ROUGE-2 and ROUGE-L are reported on the DUC2002 dataset, since the official automatic evaluation of the single document summarization task in DUC 2002 is recall based ROUGE\footnote{\url{ https://www-nlpir.nist.gov/projects/duc/guidelines/2002.html}}. 
We compute ROUGE scores using the {\tt ROUGE-1.5.5.pl} script.

Additionally, we also evaluated the generated summaries by eliciting human judgments. Following \cite{cheng:2016:acl,Narayan:ea:2018}, we randomly sampled 20 documents from the CNNDM test set. Participants were presented with a document and a list of summaries produced by different systems. We asked subjects to rank these summaries (ties allowed) by taking {informativeness} (is the summary capture the important information from the document?) and \mbox{fluency} (is the summary grammatical?) into account. Each document is annotated by three different subjects. 
% {\bf TODO:} List the models you would like to compare in human evaluation.

\subsection{Results}

\begin{table}[t]
	\centering
	% \small
	\begin{tabular}[t]{|@{~}l @{~}|@{~}c c c@{~}|}
		\hline
		Model & R-1 & R-2 & R-L \\
		\hline
		\hline
		% {\sc Lead3} \cite{Narayan:ea:2018} & 39.60 & 17.70 & 36.20 \\
		{Pointer+Coverage} & 39.53 & 17.28 & 36.38 \\
		{Abstract-ML+RL} & 39.87 & 15.82 & 36.90 \\ 
		DCA & 41.69 & 19.47 & 37.92 \\
		SentRewrite & 40.88 &  17.80 & 38.54 \\
		InconsisLoss & 40.68 & 17.97 & 37.13 \\
		Bottom-Up & 41.22 & 18.68 & 38.34 \\
		\hline\hline
		{Lead3} & 40.34 & 17.70 & 36.57 \\
		{SummaRuNNer} & 39.60 & 16.20 & 35.30 \\
		NeuSum & 40.11 & 17.52 & 36.39 \\
		{Refresh} & 40.00 & 18.20 & 36.60 \\
		NeuSum-MMR & 41.59 & 19.01 & 37.98 \\
		BanditSum & 41.50 & 18.70 & 37.60 \\
		JECS & 41.70 & 18.50 & 37.90 \\
		{LatentSum} & 41.05 & {18.77} & 37.54 \\
		HierTransformer & 41.11 & 18.69 & 37.53 \\
		BERT & 41.82 & 19.48 & 38.30 \\
		$\text{\sc Hibert}_S$ (in-domain) & 42.10 & 19.70 & 38.53 \\
		$\text{\sc Hibert}_S$ & {\bf 42.31} & {\bf 19.87} & {\bf 38.78} \\
		$\text{\sc Hibert}_M$ & {\bf 42.37} & {\bf 19.95} & {\bf 38.83} \\
		\hline
	\end{tabular}
	\caption{Results of various models on the CNNDM test
		set using full-length F1 \mbox{\textsc{Rouge-1}} (R-1),
		\textsc{Rouge-2} (R-2), and \textsc{Rouge-L} (R-L).}
	\label{tbl:cnndaily}
\end{table}

Our main results on the CNNDM dataset are shown in Table \ref{tbl:cnndaily}, with abstractive models in the top block and extractive models in the bottom block. Pointer+Coverage \cite{see:2017:acl}, Abstract-ML+RL \cite{paulus:2017:arxiv} and DCA \cite{celikyilmaz:2016:naacl} are all sequence to sequence learning based models with copy and coverage modeling, reinforcement learning and deep communicating agents extensions. SentRewrite \cite{hsu:2018:acl} and \mbox{InconsisLoss} \cite{chen:2018:acl} all try to decompose the word by word summary generation into sentence selection from document and ``sentence'' level summarization (or compression). \mbox{Bottom-Up} \cite{gehrmann:2018:emnlp} generates summaries by combines a word prediction model with the decoder attention model. The extractive models are usually based on hierarchical encoders (SummaRuNNer; \citealt{nallapati:2017:aaai} and NeuSum; \citealt{cheng:2016:acl}). They have been extended with reinforcement learning (Refresh; \citealt{Narayan:ea:2018} and BanditSum; \citealt{dong:2018:emnlp}), Maximal Marginal Relevance (NeuSum-MMR; \citealt{zhou:2018:acl}), latent variable modeling (LatentSum; \citealt{zhang:2018:emnlp}) and syntactic compression (JECS; \citealt{xu:2019:arxiv}). Lead3 is a baseline which simply selects the first three sentences. Our model $\text{\sc Hibert}_S$ (in-domain), which only use one pre-training stage on the in-domain CNNDM training set, outperforms all of them and differences between them are all significant with a 0.95 confidence interval (estimated with the ROUGE script).
% (ROUGE scores have a 95\% confidence interval of at most $\pm$0.25 as reported by the ROUGE script). 
Note that pre-training $\text{\sc Hibert}_S$ (in-domain) is very fast and it only takes around 30 minutes for one epoch on the CNNDM training set. Our models with two pre-training stages ($\text{\sc Hibert}_S$) or larger size ($\text{\sc Hibert}_M$) perform even better and $\text{\sc Hibert}_M$ outperforms BERT by 0.5 ROUGE\footnote{The difference is significant according to the ROUGE script.}. We also implemented two baselines. One is the hierarchical transformer summarization model (HeriTransfomer; described in \ref{sec:sum}) without pre-training. Note the setting for HeriTransfomer is ($L=4$,$H=300$ and $A=4$) \footnote{We tried deeper and larger models, but obtained inferior results, which may indicates training large or deep models on this dataset without a good initialization is challenging.}. We can see that the pre-training (details in Section \ref{sec:pretrain}) leads to a +1.25 ROUGE improvement. Another baseline is based on a pre-trained BERT \cite{devlin:2018:arxiv}\footnote{Our BERT baseline is adapted from this implementation \url{https://github.com/huggingface/pytorch-pretrained-BERT}} and finetuned on the CNNDM dataset. We used the $\text{BERT}_{\text{base}}$ model because our 16G RAM V100 GPU cannot fit $\text{BERT}_{\text{large}}$ for the summarization task even with batch size of 1. The positional embedding of BERT supports input length up to 512 words, we therefore split documents with more than 10 sentences into multiple blocks (each block with 10 sentences\footnote{We use 10 sentences per block, because maximum sentence length $50 \times 10 < 512$ (maximum BERT supported length). The last block of a document may have less than 10 sentences.}). We feed each block (the {\tt BOS} and {\tt EOS} tokens of each sentence are replaced with {\tt [CLS]} and {\tt [SEP]} tokens) into BERT and use the representation at {\tt [CLS]} token to classify each sentence. Our model $\text{\sc Hibert}_S$ outperforms BERT by 0.4 to 0.5 ROUGE despite with only half the number of model parameters ($\text{\sc Hibert}_S$ 54.6M v.s. BERT 110M). % put a table here?

Results on the NYT50 dataset show the similar trends (see Table \ref{tbl:nyt50}). EXTRACTION is a extractive model based hierarchical LSTM and we use the numbers reported by \newcite{xu:2019:arxiv}. The improvement of $\text{\sc Hibert}_M$ over the baseline without pre-training (HeriTransformer) becomes 2.0 ROUGE. $\text{\sc Hibert}_S$ (in-domain), $\text{\sc Hibert}_M$ (in-domain), $\text{\sc Hibert}_S$ and $\text{\sc Hibert}_M$ all outperform BERT significantly according to the ROUGE script.

\begin{table}[t]
	\centering
	% \small
	\begin{tabular}[t]{|@{~}l @{~}|@{~}c c c@{~}|}
		\hline
		Models & R-1 & R-2 & R-L \\
		\hline
		\hline
		Lead & 41.80 & 22.60 & 35.00 \\
		EXTRACTION & 44.30 & 25.50 & 37.10 \\
		JECS & 45.50 & 25.30 & 38.20 \\
		
		HeriTransformer & 47.44 & 28.08 & 39.56 \\	
		BERT & 48.38 & 29.04 & 40.53 \\ 
		$\text{\sc Hibert}_S$   (in-domain) & 48.92 & 29.58 & 41.10 \\
		$\text{\sc Hibert}_M$  (in-domain)  & 49.06 & 29.70 & 41.23 \\
		$\text{\sc Hibert}_S$  & {\bf 49.25} & {\bf 29.92} & {\bf 41.43} \\
		$\text{\sc Hibert}_M$  & {\bf 49.47} & {\bf 30.11} & {\bf 41.63} \\
		\hline
	\end{tabular}
	\caption{Results of various models on the NYT50 test set
		using full-length F1 ROUGE. $\text{\sc Hibert}_S$   (in-domain) and $\text{\sc Hibert}_M$  (in-domain) only uses one pre-training stage on the NYT50 training set.}
	\label{tbl:nyt50}
\end{table}

\begin{table}[t]
	\centering
	% \small
	\begin{tabular}[t]{|@{~}l @{~}|@{~}c c c@{~}|}
		\hline
		Pretraining Strategies & R-1 & R-2 & R-L \\
		\hline
		\hline
		% {\sc Lead3} \cite{Narayan:ea:2018} & 39.60 & 17.70 & 36.20 \\
		Open-Domain & 42.97 & 20.31 & 39.51 \\ 
		In-Domain & 42.93 & 20.28 & 39.46 \\
		Open+In-Domain & {\bf 43.19} & {\bf 20.46} & {\bf 39.72} \\
		\hline
	\end{tabular}
	\caption{Results of summarization model ($\text{\sc Hibert}_S$ setting) with different pre-training strategies on the CNNDM validation set using full-length F1 ROUGE.}
	\label{tbl:pretrain}
\end{table}

\begin{table}[t]
	\centering
	\small
	\setlength\tabcolsep{3.5pt} 
	\begin{tabular}[t]{| l | c c c c c c | c |}
		\hline
		Models & 1st & 2nd & 3rd & 4th & 5th & 6th & MeanR \\
		\hline
		\hline
		Lead3 & 0.03 & 0.18 & 0.15 & 0.30 & 0.30 & 0.03 & 3.75 \\
		DCA & 0.08 & 0.15 & 0.18 & 0.20 & 0.15 & 0.23 & 3.88 \\ 
		Latent & 0.05 & 0.33 & 0.28 & 0.20 & 0.13 & 0.00 & 3.03 \\
		BERT & 0.13 & 0.37 & 0.32 & 0.15 &  0.03 & 0.00 & 2.58 \\
		$\text{\sc Hibert}_M$  & 0.30  & 0.35  & 0.25 & 0.10 & 0.00 & 0.00 & 2.15  \\
		Human  & 0.58 & 0.15 & 0.20 & 0.00 & 0.03 & 0.03 & 1.85 \\
		\hline
	\end{tabular}
	\caption{Human evaluation: proportions of rankings and mean ranks (MeanR; lower is better) of various models.}
	\label{tbl:human}
\end{table}

We also conducted human experiment with 20 randomly sampled documents from the CNNDM test set. We compared our model $\text{\sc Hibert}_M$ against Lead3, DCA, Latent, BERT and the human reference (Human)\footnote{We obtained the outputs of DCA and Latent via emails.}. We asked the subjects to rank the outputs of these systems from best to worst. As shown in Table \ref{tbl:human}, the output of $\text{\sc Hibert}_M$ is selected as the best in 30\% of cases and we obtained lower mean rank than all systems except for \mbox{Human}. We also converted the rank numbers into ratings (rank $i$ to $7-i$) and applied student $t$-test on the ratings. $\text{\sc Hibert}_M$ is significantly different from all systems in comparison ($p < 0.05$), which indicates our model still lags behind Human, but is better than all other systems.

\paragraph{Pre-training Strategies}
As mentioned earlier, our pre-training includes two stages. The first stage is the open-domain pre-training stage on the GIGA-CM dataset and the following stage is the in-domain pre-training on the \mbox{CNNDM} (or NYT50) dataset. As shown in Table \ref{tbl:pretrain}, we pretrained $\text{\sc Hibert}_S$ using only open-domain stage (Open-Domain), only in-domain stage (In-Domain) or both stages (Open+In-Domain) and applied it to the CNNDM summarization task. Results on the validation set of CNNDM indicate the two-stage pre-training process is necessary.

%
%\begin{table}[t]
%	\centering
%	% \small
%	\begin{tabular}[t]{|@{~}l @{~}|@{~}c c c@{~}|}
%		\hline
%		Pretraining Strategies & R-1 & R-2 & R-L \\
%		\hline
%		\hline
%		% {\sc Lead3} \cite{Narayan:ea:2018} & 39.60 & 17.70 & 36.20 \\
%		In-Domain & 42.08 & 19.70 & 38.53 \\
%		Open-Domain & 41.74 & 19.43 & 38.18 \\ 
%		Open+In-Domain & {\bf 42.31} & {\bf 19.87} & {\bf 38.78} \\
%		\hline
%	\end{tabular}
%	\caption{Results of summarization models with different pre-training strategies on the CNNDM validation set using full-length F1 ROUGE.}
%	
%	\label{tbl:pretrain}
%\end{table}

\section{Conclusions}
The core part of a neural extractive summarization model is the hierarchical document encoder. 
We proposed a method to pre-train document level hierarchical bidirectional transformer encoders on unlabeled data. When we only pre-train hierarchical transformers on the training sets of summarization datasets with our proposed objective, application of the pre-trained hierarchical transformers to extractive summarization models already leads to wide improvement of summarization performance. Adding the large open-domain dataset to pre-training leads to even better performance. In the future, we plan to apply models to other tasks that also require hierarchical document encodings (e.g., document question answering). We are also interested in improving the architectures of hierarchical document encoders and designing other objectives to train hierarchical transformers.

\bibliography{acl2019}

\begin{thebibliography}{45}
\expandafter\ifx\csname natexlab\endcsname\relax\def\natexlab#1{#1}\fi

\bibitem[{Ba et~al.(2016)Ba, Kiros, and Hinton}]{ba:2016:arxiv}
Jimmy~Lei Ba, Jamie~Ryan Kiros, and Geoffrey~E Hinton. 2016.
\newblock Layer normalization.
\newblock \emph{arXiv preprint arXiv:1607.06450}.

\bibitem[{Bahdanau et~al.(2015)Bahdanau, Cho, and Bengio}]{bahdanau:2015:iclr}
Dzmitry Bahdanau, Kyunghyun Cho, and Yoshua Bengio. 2015.
\newblock Neural machine translation by jointly learning to align and
  translate.
\newblock In \emph{In Proceedings of the 3rd International Conference on
  Learning Representations}, San Diego, California.

\bibitem[{Bojanowski et~al.(2017)Bojanowski, Grave, Joulin, and
  Mikolov}]{bojanowski:2017:tacl}
Piotr Bojanowski, Edouard Grave, Armand Joulin, and Tomas Mikolov. 2017.
\newblock Enriching word vectors with subword information.
\newblock \emph{Transactions of the Association for Computational Linguistics},
  5:135--146.

\bibitem[{Celikyilmaz et~al.(2018)Celikyilmaz, Bosselut, He, and
  Choi}]{celikyilmaz:2016:naacl}
Asli Celikyilmaz, Antoine Bosselut, Xiaodong He, and Yejin Choi. 2018.
\newblock Deep communicating agents for abstractive summarization.
\newblock In \emph{Proceedings of the 2018 Conference of the North American
  Chapter of the Association for Computational Linguistics: Human Language
  Technologies, Volume 1 (Long Papers)}, pages 1662--1675, New Orleans,
  Louisiana.

\bibitem[{Chen et~al.(2018)Chen, Gao, Tao, Song, Zhao, and
  Yan}]{chen:2018:emnlp}
Xiuying Chen, Shen Gao, Chongyang Tao, Yan Song, Dongyan Zhao, and Rui Yan.
  2018.
\newblock \href {http://aclweb.org/anthology/D18-1442} {Iterative document
  representation learning towards summarization with polishing}.
\newblock In \emph{Proceedings of the 2018 Conference on Empirical Methods in
  Natural Language Processing}, pages 4088--4097. Association for Computational
  Linguistics.

\bibitem[{Chen and Bansal(2018)}]{chen:2018:acl}
Yen-Chun Chen and Mohit Bansal. 2018.
\newblock \href {http://aclweb.org/anthology/P18-1063} {Fast abstractive
  summarization with reinforce-selected sentence rewriting}.
\newblock In \emph{Proceedings of the 56th Annual Meeting of the Association
  for Computational Linguistics (Volume 1: Long Papers)}, pages 675--686.
  Association for Computational Linguistics.

\bibitem[{Cheng and Lapata(2016)}]{cheng:2016:acl}
Jianpeng Cheng and Mirella Lapata. 2016.
\newblock Neural summarization by extracting sentences and words.
\newblock In \emph{Proceedings of the 54th Annual Meeting of the Association
  for Computational Linguistics (Volume 1: Long Papers)}, pages 484--494,
  Berlin, Germany.

\bibitem[{Conroy and O'leary(2001)}]{conroy:2001:sigir}
John~M Conroy and Dianne~P O'leary. 2001.
\newblock Text summarization via hidden markov models.
\newblock In \emph{Proceedings of the 24th annual international ACM SIGIR
  conference on Research and development in information retrieval}, pages
  406--407. ACM.

\bibitem[{Devlin et~al.(2018)Devlin, Chang, Lee, and
  Toutanova}]{devlin:2018:arxiv}
Jacob Devlin, Ming-Wei Chang, Kenton Lee, and Kristina Toutanova. 2018.
\newblock {BERT: Pre-training of Deep Bidirectional Transformers for Language
  Understanding}.
\newblock \emph{arXiv preprint arXiv:1810.04805}.

\bibitem[{Dong et~al.(2018)Dong, Shen, Crawford, van Hoof, and
  Cheung}]{dong:2018:emnlp}
Yue Dong, Yikang Shen, Eric Crawford, Herke van Hoof, and Jackie Chi~Kit
  Cheung. 2018.
\newblock \href {http://aclweb.org/anthology/D18-1409} {Banditsum: Extractive
  summarization as a contextual bandit}.
\newblock In \emph{Proceedings of the 2018 Conference on Empirical Methods in
  Natural Language Processing}, pages 3739--3748. Association for Computational
  Linguistics.

\bibitem[{Durrett et~al.(2016)Durrett, Berg-Kirkpatrick, and
  Klein}]{durrett:2016:acl}
Greg Durrett, Taylor Berg-Kirkpatrick, and Dan Klein. 2016.
\newblock \href {https://doi.org/10.18653/v1/P16-1188} {Learning-based
  single-document summarization with compression and anaphoricity constraints}.
\newblock In \emph{Proceedings of the 54th Annual Meeting of the Association
  for Computational Linguistics (Volume 1: Long Papers)}, pages 1998--2008.
  Association for Computational Linguistics.

\bibitem[{Filatova and
  Hatzivassiloglou(2004{\natexlab{a}})}]{filatova:2004:acl:workshop}
Elena Filatova and Vasileios Hatzivassiloglou. 2004{\natexlab{a}}.
\newblock Event-based extractive summarization.
\newblock In \emph{Text Summarization Branches Out: Proceedings of the ACL-04
  Workshop}, pages 104--111, Barcelona, Spain.

\bibitem[{Filatova and
  Hatzivassiloglou(2004{\natexlab{b}})}]{filatova2004event}
Elena Filatova and Vasileios Hatzivassiloglou. 2004{\natexlab{b}}.
\newblock Event-based extractive summarization.

\bibitem[{Gehrmann et~al.(2018)Gehrmann, Deng, and Rush}]{gehrmann:2018:emnlp}
Sebastian Gehrmann, Yuntian Deng, and Alexander Rush. 2018.
\newblock \href {http://aclweb.org/anthology/D18-1443} {Bottom-up abstractive
  summarization}.
\newblock In \emph{Proceedings of the 2018 Conference on Empirical Methods in
  Natural Language Processing}, pages 4098--4109. Association for Computational
  Linguistics.

\bibitem[{Glorot et~al.(2011)Glorot, Bordes, and Bengio}]{glorot:2011:aistats}
Xavier Glorot, Antoine Bordes, and Yoshua Bengio. 2011.
\newblock Deep sparse rectifier neural networks.
\newblock In \emph{Proceedings of the fourteenth international conference on
  artificial intelligence and statistics}, pages 315--323.

\bibitem[{Gu et~al.(2016)Gu, Lu, Li, and Li}]{gu:2016:acl}
Jiatao Gu, Zhengdong Lu, Hang Li, and Victor~O.K. Li. 2016.
\newblock \href {https://doi.org/10.18653/v1/P16-1154} {Incorporating copying
  mechanism in sequence-to-sequence learning}.
\newblock In \emph{Proceedings of the 54th Annual Meeting of the Association
  for Computational Linguistics (Volume 1: Long Papers)}, pages 1631--1640.
  Association for Computational Linguistics.

\bibitem[{He et~al.(2016)He, Zhang, Ren, and Sun}]{he:2016:cvpr}
Kaiming He, Xiangyu Zhang, Shaoqing Ren, and Jian Sun. 2016.
\newblock Deep residual learning for image recognition.
\newblock In \emph{Proceedings of the IEEE conference on computer vision and
  pattern recognition}, pages 770--778.

\bibitem[{Hermann et~al.(2015)Hermann, Kocisky, Grefenstette, Espeholt, Kay,
  Suleyman, and Blunsom}]{hermann:2015:nips}
Karl~Moritz Hermann, Tomas Kocisky, Edward Grefenstette, Lasse Espeholt, Will
  Kay, Mustafa Suleyman, and Phil Blunsom. 2015.
\newblock Teaching machines to read and comprehend.
\newblock In \emph{Advances in Neural Information Processing Systems}, pages
  1693--1701. Curran Associates, Inc.

\bibitem[{Hochreiter and Schmidhuber(1997)}]{hochreiter:1997:nc}
Sepp Hochreiter and J{\"u}rgen Schmidhuber. 1997.
\newblock Long short-term memory.
\newblock \emph{Neural computation}, 9(8):1735--1780.

\bibitem[{Hsu et~al.(2018)Hsu, Lin, Lee, Min, Tang, and Sun}]{hsu:2018:acl}
Wan-Ting Hsu, Chieh-Kai Lin, Ming-Ying Lee, Kerui Min, Jing Tang, and Min Sun.
  2018.
\newblock \href {http://aclweb.org/anthology/P18-1013} {A unified model for
  extractive and abstractive summarization using inconsistency loss}.
\newblock In \emph{Proceedings of the 56th Annual Meeting of the Association
  for Computational Linguistics (Volume 1: Long Papers)}, pages 132--141.
  Association for Computational Linguistics.

\bibitem[{Kupiec et~al.(1995)Kupiec, Pedersen, and Chen}]{kupiec:1995:sigir}
Julian Kupiec, Jan Pedersen, and Francine Chen. 1995.
\newblock A trainable document summarizer.
\newblock In \emph{Proceedings of the 18th annual international ACM SIGIR
  conference on Research and development in information retrieval}, pages
  68--73. ACM.

\bibitem[{Lin(2004)}]{lin:2004:acl:w}
Chin-Yew Lin. 2004.
\newblock Rouge: A package for automatic evaluation of summaries.
\newblock In \emph{Text Summarization Branches Out: Proceedings of the ACL-04
  Workshop}, pages 74--81, Barcelona, Spain.

\bibitem[{Mani(2001)}]{Mani:01}
Inderjeet Mani. 2001.
\newblock \emph{Automatic Summarization}.
\newblock John Benjamins Pub Co.

\bibitem[{Manning et~al.(2014)Manning, Surdeanu, Bauer, Finkel, Bethard, and
  McClosky}]{manning:2014:acldemo}
Christopher Manning, Mihai Surdeanu, John Bauer, Jenny Finkel, Steven Bethard,
  and David McClosky. 2014.
\newblock The stanford corenlp natural language processing toolkit.
\newblock In \emph{Proceedings of 52nd annual meeting of the association for
  computational linguistics: system demonstrations}, pages 55--60.

\bibitem[{Mikolov et~al.(2013)Mikolov, Sutskever, Chen, Corrado, and
  Dean}]{mikolov:2013:nips}
Tomas Mikolov, Ilya Sutskever, Kai Chen, Greg~S Corrado, and Jeff Dean. 2013.
\newblock Distributed representations of words and phrases and their
  compositionality.
\newblock In \emph{Advances in neural information processing systems}, pages
  3111--3119.

\bibitem[{Nallapati et~al.(2017)Nallapati, Zhai, and
  Zhou}]{nallapati:2017:aaai}
Ramesh Nallapati, Feifei Zhai, and Bowen Zhou. 2017.
\newblock Summarunner: A recurrent neural network based sequence model for
  extractive summarization of documents.
\newblock In \emph{In Proceedings of the 31st AAAI Conference on Artificial
  Intelligence}, pages 3075--3091, San Francisco, California.

\bibitem[{Nallapati et~al.(2016)Nallapati, Zhou, Gulcehre, Xiang
  et~al.}]{nallapati:2016:arxiv}
Ramesh Nallapati, Bowen Zhou, Caglar Gulcehre, Bing Xiang, et~al. 2016.
\newblock Abstractive text summarization using sequence-to-sequence rnns and
  beyond.
\newblock \emph{arXiv preprint arXiv:1602.06023}.

\bibitem[{Narayan et~al.(2018)Narayan, Cohen, and Lapata}]{Narayan:ea:2018}
Shashi Narayan, Shay~B. Cohen, and Mirella Lapata. 2018.
\newblock Ranking sentences for extractive summarization with reinforcement
  learning.
\newblock In \emph{Proceedings of the 2018 Conference of the North American
  Chapter of the Association for Computational Linguistics: Human Language
  Technologies, Volume 1 (Long Papers)}, pages 1747--1759, New Orleans,
  Louisiana.

\bibitem[{Nenkova and McKeown(2011)}]{Nenkova:McKeown:2011}
Ani Nenkova and Kathleen McKeown. 2011.
\newblock Automatic summarization.
\newblock \emph{Foundations and Trends in Information Retrieval},
  5(2--3):103--233.

\bibitem[{Nenkova et~al.(2006)Nenkova, Vanderwende, and
  McKeown}]{nenkova:2006:sigir}
Ani Nenkova, Lucy Vanderwende, and Kathleen McKeown. 2006.
\newblock A compositional context sensitive multi-document summarizer:
  exploring the factors that influence summarization.
\newblock In \emph{Proceedings of the 29th annual international ACM SIGIR
  conference on Research and development in information retrieval}, pages
  573--580. ACM.

\bibitem[{Paulus et~al.(2017)Paulus, Xiong, and Socher}]{paulus:2017:arxiv}
Romain Paulus, Caiming Xiong, and Richard Socher. 2017.
\newblock A deep reinforced model for abstractive summarization.
\newblock \emph{arXiv preprint arXiv:1705.04304}.

\bibitem[{Pennington et~al.(2014)Pennington, Socher, and
  Manning}]{pennington:2014:emnlp}
Jeffrey Pennington, Richard Socher, and Christopher Manning. 2014.
\newblock Glove: Global vectors for word representation.
\newblock In \emph{Proceedings of the 2014 conference on empirical methods in
  natural language processing (EMNLP)}, pages 1532--1543.

\bibitem[{Peters et~al.(2018)Peters, Neumann, Iyyer, Gardner, Clark, Lee, and
  Zettlemoyer}]{peters:2018:naacl}
Matthew Peters, Mark Neumann, Mohit Iyyer, Matt Gardner, Christopher Clark,
  Kenton Lee, and Luke Zettlemoyer. 2018.
\newblock \href {https://doi.org/10.18653/v1/N18-1202} {Deep contextualized
  word representations}.
\newblock In \emph{Proceedings of the 2018 Conference of the North American
  Chapter of the Association for Computational Linguistics: Human Language
  Technologies, Volume 1 (Long Papers)}, pages 2227--2237. Association for
  Computational Linguistics.

\bibitem[{Radev et~al.(2004)Radev, Allison, Blair-Goldensohn, Blitzer,
  {\c{C}}elebi, Dimitrov, Drabek, Hakim, Lam, Liu, Otterbacher, Qi, Saggion,
  Teufel, Topper, Winkel, and Zhang}]{radev:2004}
Dragomir Radev, Timothy Allison, Sasha Blair-Goldensohn, John Blitzer, Arda
  {\c{C}}elebi, Stanko Dimitrov, Elliott Drabek, Ali Hakim, Wai Lam, Danyu Liu,
  Jahna Otterbacher, Hong Qi, Horacio Saggion, Simone Teufel, Michael Topper,
  Adam Winkel, and Zhu Zhang. 2004.
\newblock \href {http://www.lrec-conf.org/proceedings/lrec2004/pdf/757.pdf}
  {Mead - a platform for multidocument multilingual text summarization}.
\newblock In \emph{Proceedings of the Fourth International Conference on
  Language Resources and Evaluation (LREC'04)}. European Language Resources
  Association (ELRA).

\bibitem[{Radford et~al.(2018)Radford, Narasimhan, Salimans, and
  Sutskever}]{radford:2018:nips}
Alec Radford, Karthik Narasimhan, Tim Salimans, and Ilya Sutskever. 2018.
\newblock Improving language understanding by generative pre-training.
\newblock \emph{URL https://s3-us-west-2. amazonaws.
  com/openai-assets/research-covers/languageunsupervised/language understanding
  paper. pdf}.

\bibitem[{See et~al.(2017)See, Liu, and Manning}]{see:2017:acl}
Abigail See, Peter~J. Liu, and Christopher~D. Manning. 2017.
\newblock \href {http://aclweb.org/anthology/P17-1099} {Get to the point:
  Summarization with pointer-generator networks}.
\newblock In \emph{Proceedings of the 55th Annual Meeting of the Association
  for Computational Linguistics (Volume 1: Long Papers)}, pages 1073--1083,
  Vancouver, Canada.

\bibitem[{Sennrich et~al.(2016)Sennrich, Haddow, and Birch}]{sennrich:2016:acl}
Rico Sennrich, Barry Haddow, and Alexandra Birch. 2016.
\newblock \href {https://doi.org/10.18653/v1/P16-1162} {Neural machine
  translation of rare words with subword units}.
\newblock In \emph{Proceedings of the 54th Annual Meeting of the Association
  for Computational Linguistics (Volume 1: Long Papers)}, pages 1715--1725.
  Association for Computational Linguistics.

\bibitem[{Strubell et~al.(2018)Strubell, Verga, Andor, Weiss, and
  McCallum}]{strubell:2018:emnlp}
Emma Strubell, Patrick Verga, Daniel Andor, David Weiss, and Andrew McCallum.
  2018.
\newblock \href {http://aclweb.org/anthology/D18-1548} {Linguistically-informed
  self-attention for semantic role labeling}.
\newblock In \emph{Proceedings of the 2018 Conference on Empirical Methods in
  Natural Language Processing}, pages 5027--5038. Association for Computational
  Linguistics.

\bibitem[{Sutskever et~al.(2014)Sutskever, Vinyals, and
  Le}]{sutskever:2014:nips}
Ilya Sutskever, Oriol Vinyals, and Quoc~V Le. 2014.
\newblock Sequence to sequence learning with neural networks.
\newblock In \emph{Advances in neural information processing systems}, pages
  3104--3112.

\bibitem[{Taylor(1953)}]{taylor:1953:sage}
Wilson~L Taylor. 1953.
\newblock “cloze procedure”: A new tool for measuring readability.
\newblock \emph{Journalism Bulletin}, 30(4):415--433.

\bibitem[{Vaswani et~al.(2017)Vaswani, Shazeer, Parmar, Uszkoreit, Jones,
  Gomez, Kaiser, and Polosukhin}]{vaswani:2017:nips}
Ashish Vaswani, Noam Shazeer, Niki Parmar, Jakob Uszkoreit, Llion Jones,
  Aidan~N Gomez, {\L}ukasz Kaiser, and Illia Polosukhin. 2017.
\newblock Attention is all you need.
\newblock In \emph{Advances in Neural Information Processing Systems}, pages
  5998--6008.

\bibitem[{Woodsend and Lapata(2010)}]{woodsend:2010:acl}
Kristian Woodsend and Mirella Lapata. 2010.
\newblock Automatic generation of story highlights.
\newblock In \emph{Proceedings of the 48th Annual Meeting of the Association
  for Computational Linguistics}, pages 565--574, Uppsala, Sweden.

\bibitem[{Xu and Durrett(2019)}]{xu:2019:arxiv}
Jiacheng Xu and Greg Durrett. 2019.
\newblock Neural extractive text summarization with syntactic compression.
\newblock \emph{arXiv preprint arXiv:1902.00863}.

\bibitem[{Zhang et~al.(2018)Zhang, Lapata, Wei, and Zhou}]{zhang:2018:emnlp}
Xingxing Zhang, Mirella Lapata, Furu Wei, and Ming Zhou. 2018.
\newblock \href {http://aclweb.org/anthology/D18-1088} {Neural latent
  extractive document summarization}.
\newblock In \emph{Proceedings of the 2018 Conference on Empirical Methods in
  Natural Language Processing}, pages 779--784. Association for Computational
  Linguistics.

\bibitem[{Zhou et~al.(2018)Zhou, Yang, Wei, Huang, Zhou, and
  Zhao}]{zhou:2018:acl}
Qingyu Zhou, Nan Yang, Furu Wei, Shaohan Huang, Ming Zhou, and Tiejun Zhao.
  2018.
\newblock \href {http://aclweb.org/anthology/P18-1061} {Neural document
  summarization by jointly learning to score and select sentences}.
\newblock In \emph{Proceedings of the 56th Annual Meeting of the Association
  for Computational Linguistics (Volume 1: Long Papers)}, pages 654--663.
  Association for Computational Linguistics.

\end{thebibliography}
\bibliographystyle{acl_natbib}

\end{document}